\documentclass[journal]{IEEEtran}

\usepackage{times}
\usepackage{epsfig}
\usepackage{graphicx}
\usepackage{amsmath}
\usepackage{amssymb}

\usepackage{subfigure}

\usepackage{algorithm}  
\usepackage{algorithmicx}  
\usepackage{algpseudocode}  
\usepackage{booktabs}
\usepackage{multirow}
\usepackage{color}

\ifCLASSINFOpdf
\else
\fi
\begin{document}
%
\title{$ \rm \textbf{U}^2$-ONet$ $: A Two-level Nested Octave U-structure with Multiscale Attention Mechanism for Moving Instances Segmentation}
%
%
%

\author{Chenjie Wang$^{1}$ and Chengyuan Li$^{1}$ and Bin Luo$^{1}$
	\thanks{$^{1}$State Key Laboratory of Information Engineering in Surveying, Mapping and Remote Sensing, Wuhan University, Wuhan, Hubei, China
		{\tt\small \{wangchenjie, lichengyuan, luob\}@whu.edu.cn}}%
}

\maketitle

\begin{abstract}
Most scenes in practical applications are dynamic scenes containing moving objects, so segmenting accurately moving objects is crucial for many computer vision applications. In order to efficiently segment out all moving objects in the scene, regardless of whether the object has a predefined semantic label, we propose a two-level nested Octave U-structure network with a multiscale attention mechanism called $ \rm \textbf{U}^2$-ONe$ \rm \textbf{t}$. Each stage of $ \rm \textbf{U}^2$-ONe$ \rm \textbf{t}$ is filled with our newly designed Octave ReSidual U-block (ORSU) to enhance the ability to obtain more context information at different scales while reducing spatial redundancy of feature maps. In order to efficiently train our multi-scale deep network, we introduce a hierarchical training supervision strategy that calculates the loss at each level while adding a knowledge matching loss to keep the optimization consistency. Experimental results show that our method achieves state-of-the-art performance in several general moving objects segmentation datasets.
\end{abstract}


%
\IEEEpeerreviewmaketitle

\section{Introduction}

Moving object instances segmentation is a very critical technology in computer vision tasks, which is directly related to the effects of subsequent work such as object tracking, visual SLAM (simultaneous localization and mapping), image recognition, etc. Being able to accurately segment moving objects from a video sequence can greatly improve the effects of many tasks in dynamic scenes such as dynamic visual SLAM~\cite{saputra2018visual}, dynamic object obstacle avoidance, and dynamic object modeling, etc. For example, in an autonomous driving scene, the segmentation of moving objects can help understand surrounding motion information, which is the basis for avoiding collision, braking operations and smooth maneuvering. Most of the current methods are designed to segment N pre-defined classes in the training set. However, in practical environment, many applications such as autonomous driving and intelligent robots need to achieve robust perception in the open-world. They all must discover and segment never-before-seen moving objects in the new environment, regardless of whether it is associated with a particular semantic class. 

The segmentation of different motions in dynamic scenes has been studied for decades. Traditional methods of motion segmentation use powerful geometric constraints to cluster the points in the scene into a model parameter instance, thereby segmenting moving objects in different motions~\cite{zhao2019motion,zhang2017permutation}, which called multi-motion segmentation. This kind of methods realizes the motion segmentation of feature points instead of pixel-by-pixel. And these methods are not enough robust and versatile, because they can only segment the more salient moving objects in the scene, and the number of motion models that can be segmented at once is limited. With the development of deep learning, instance/semantic segmentation and object detection in videos has been well studied~\cite{he2017mask,badrinarayanan2017segnet,ronneberger2015u,redmon2018yolov3}. These methods are used to segment specific labeled object categories in annotated data, so the main focus is on pre-defined semantic category segmentation through appearance rather than all moving instances segmentation. Meanwhile, they are not able to segment new objects that have not been labeled in the training data. More recent approaches combine instance/semantic segmentation results with motion information from optical flow to segment moving object instances in dynamic scene, as in~\cite{bideau2018best,xie2019object,dave2019towards,muthu2020motion}. Among them, the methods~\cite{xie2019object,dave2019towards} can segment never-before-seen objects which has not been predefined in the training set based on their motion, like our work. However, these methods did not explore the use of the deep architecture which has been proved successful in many artificial intelligence tasks. By comparison, we use deeper architecture inspired by~\cite{qin2020u2} without sacrificing high resolution to improve the effect of moving instances segmentation. As we all known, the deeper network brings more spatial redundancy of feature maps, computation and memory cost and increases the difficulty of training supervision. Therefore, on the one hand we improve ReSidual U-block (RSU) in ~\cite{qin2020u2} with Octave Convolution (OctConv)~\cite{chen2019drop} and design a novel ORSU block. We take advantage of OctConv to reduce the spatial redundancy and further improve the accuracy while greatly reducing computation and memory resources. On the other hand, we design a hierarchical training supervision strategy to force the consistency of deep network optimization for imporving the segmentaion accuracy.

\begin{figure*}
	\begin{center}
		\includegraphics[width=0.8\linewidth]{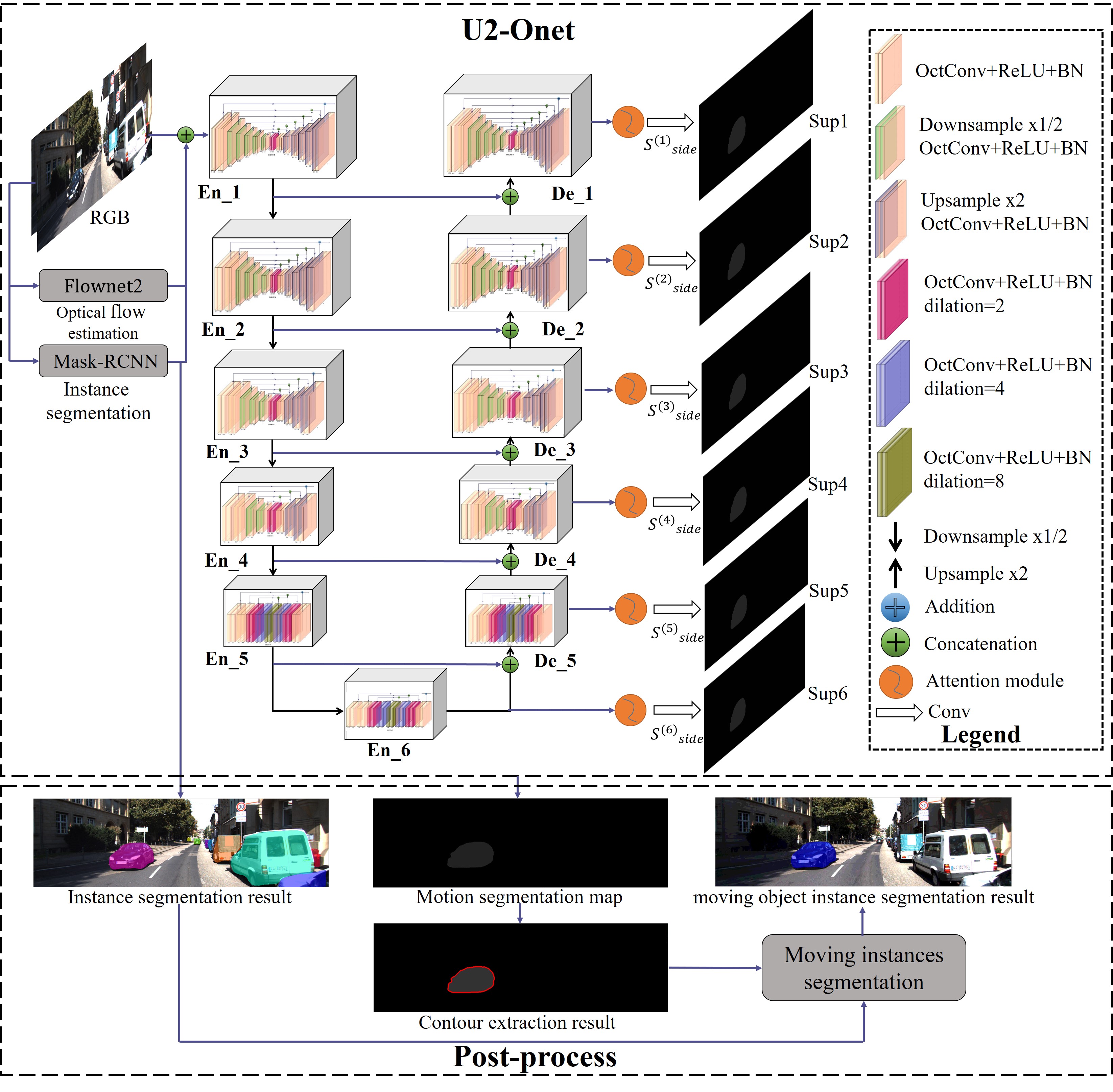}
	\end{center}
	\caption{Illustration of our $ \rm U^2$-ONe$ \rm t$.}
	\label{fig:fig1}
\end{figure*}

In this work, we design a novel two-level nested U-structure with multiscale attention mechanism called $ \rm U^2$-ONe$ \rm t$ that takes video frames, optical flow and instance segmentation results as inputs and learns to segment pixels belonging to foreground moving objects from background. Inspired by~\cite{qin2020u2}, our $ \rm U^2$-ONe$ \rm t$ is a nested U-structure that is designed without using any pre-trained backbones from image classification. We further reduce the memory and computational cost while ensuring the accuracy. On the bottom level, we design a novel Octave ReSidual U-block (ORSU) based on RSU (ReSidual U-block) in~\cite{qin2020u2}, which uses Octave Convolution (OctConv)~\cite{chen2019drop} factorizing the mixed feature maps by their frequencies, instead of the vanilla convolution for operating. With the advantages of OctConv and structure of U-blocks where most operations are applied on the downsampled feature maps, ORSU can extract intra-stage multi-scale features without degrading the feature map resolution while reducing the representation redundancy, memory cost and computation cost. On the top level, there is a U-Net like structure, in which each stage is filled by a ORSU block and each scale contains an attention block. By adding attention block at different scales, we introduce spatial and channel attention in our network and eliminate aliasing effects. In training strategy, we design a hierarchical training supervision strategy instead of using the standard top-most supervised training and the deeply supervised training scheme. We calculate the loss at each level and add a probability matching loss called Kullback-Leibler divergence (KLloss) to promote supervision interactions among different level to guarantee more robust optimization process and better representation ability. Illustration of $ \rm U^2$-ONe$ \rm t$ are shown in Fig.~\ref{fig:fig1}.

In summary, our work has the following key contributions:
\begin{enumerate}
	\renewcommand{\labelenumi}{(\theenumi)}
	\item We propose $ \rm U^2$-ONe$ \rm t$ which is a two-level nested U-structure with multiscale attention mechanism to efficiently segment out all moving object instances in dynamic scene, regardless of whether it is associated with a particular semantic class.
	\item We design a novel Octave U-block (ORSU) with Octave Convolution to fill each stage of $ \rm U^2$-ONe$ \rm t$ which extracts intra-stage multi-scale features while adopting more efficient inter frequency information exchange as well as reducing the spatial redundancy and memory cost in CNNs.
	\item We design a hierarchical training supervision strategy which calculates both the standard binary cross-entropy (BCEloss) and KLloss at each level, and uses KLloss implicit constraint gradient to enhance the opportunity of knowledge sharing and forces the optimization consistency across the whole deep network.
	\item In the task of moving instances segmentation, we prove that $ \rm U^2$-ONe$ \rm t$ is efficient and our hierarchical training supervision strategy improves the effect of our deep network. Experimental results show that our work achieves state-of-the-art performance in some challenging datasets that including camouflaged objects, tiny objects.
\end{enumerate}

\section{Related work}
\subsection{Video Foreground Segmentation}
Video foreground segmentation have focused on the classifying every pixel in a video as foreground or background. Early methods~\cite{faktor2014video,wang2015saliency} depended on the heuristics in the optical flow field, such as spatial edges and temporal motion boundaries in~\cite{wang2015saliency} to identified video foreground objects. With the introduction of a standard benchmark, DAVIS 2016~\cite{perazzi2016benchmark}, video object segmentation has done many related researches~\cite{wang2019learning,wang2019zero,lu2019see,yang2019anchor,zhuo2019unsupervised,yang2020collaborative}. Methods~\cite{wang2019learning,lu2019see,yang2019anchor,zhuo2019unsupervised} only complete segmentation of the foreground objects and the background without segmenting individual instances. In instance-level methods, AGNN~\cite{wang2019zero} is based on a novel attentive graph neural network, and CFBI~\cite{yang2020collaborative} imposes the feature embedding from both foreground and background to perform the matching process from both pixel and instance levels. However, video object segmentation is usually to segment the most salient and critical object in videos, not just segmenting moving objects. Our method focus on motion information and is used to segment out all moving objects in the scene regardless of whether they are salient.

\subsection{Instance segmentation}
Instance segmentation not only needs to assign class labels to pixels, but also segment individual object instances in images. Methods based on R-CNN~\cite{girshick2014rich} are popular and widely used in the present time. Mask-RCNN~\cite{he2017mask} uses the object bounding box obtained by Faster R-CNN~\cite{ren2015faster} to distinguish each instance, and then segment the instances in each bounding box. PANet~\cite{liu2018path} improves Mask R-CNN by adding bottom-up path augmentation that enhances the entire feature hierarchy with accurate localization signals in earlier layers of the network. Subsequently, BshapeNet~\cite{kang2020bshapenet} designs extended frameworks by adding a bounding box mask branch providing additional information of object positions and coordinates to a Faster R-CNN to enhance the performance of instance segmentation. Recently, a few works propose some novel contour-based approaches for real-time instance segmentation~\cite{peng2020deep,hurtik2020poly}. Deep Snake~\cite{peng2020deep} uses the circular convolution for feature learning on the contour and proposes a two-stage pipeline including initial contour proposal and contour deformation for instance segmentation. POLY-YOLO~\cite{hurtik2020poly} increases the detection accuracy of YOLOv3 and realizes instance segmentation using tight polygon-based contour. However, most of these methods focus on appearance cues and can only segment objects that have been labeled as a specific category in a training set. We leverage the semantic instance masks from Mask-RCNN with motion cues from optical flow to segment moving object instances whether it is associated with a particular semantic category.  

\subsection{Motion Segmentation}
The multi-motion segmentation method~\cite{zhao2019motion,zhang2017permutation,xu20193d,thakoor2010multibody} based on the geometric involves clustering points of the same motion into a motion model parameter instance to segment the multiple motion models of the scene which can be utilized to discover new objects based on their motion. This kind of method realizes the result of feature points level instead of pixel-by-pixel, and the application conditions and scenarios are limited because of only segmenting more salient moving object, limited model numbers of segmentation and slow computational speed. More recent approaches propose optical flow based method for motion segmentation, as in~\cite{bideau2018best,xie2019object,dave2019towards,muthu2020motion,ranjan2019competitive,pan2019joint,shen2018submodular}. CC~\cite{ranjan2019competitive} introduces a generic framework called Competitive Collaboration (CC), in which four networks including motion segmentation network learn to collaborate and compete, thereby achieving specific goals such as segmenting the scene into moving objects and the static background. The method~\cite{pan2019joint} takes advantage of the interconnectedness of three tasks including moving object segmentation and solves them jointly in a unified framework. However, both methods cannot segment each object instance. Instance-level segmentation methods include designing a hierarchical motion segmentation system which combines geometric knowledge with a modern CNN for appearance modeling~\cite{bideau2018best}, introducing a novel pixel-trajectory recurrent neural network to cluster foreground pixels in videos into different objects~\cite{xie2019object}, using a two-stream architecture to separately process motion and appearance~\cite{dave2019towards}, designing a new submodular optimization process to achieve trajectory clustering~\cite{shen2018submodular} and using statistical inference for the combination of motion and semantic cues~\cite{muthu2020motion}. By comparison, we propose a two-level nested U-structure deep network with Octave Convolution to segment each moving object instances while reducing the spatial redundancy and memory cost in CNNs.

\section{Method}
First, we introduce the overall structure of the network, including the network inputs. Next, we introduce the design of our proposed Octave ReSidual U-block (ORSU-blocks) and then describe the details of the nested U-Net like structure built with ORSU-blocks and multiscale attention mechanism. Then, the hierarchical training supervision strategy we designed and the training loss are described. Post-processing to obtain instance-level moving object segmentation results are introduced at the end of this section.

\subsection{Overrall}   
Our approach takes video frames, instance segmentation of frames, optical flow between pairs of frames as inputs, which are concatenated in the channel dimension and fed through $ \rm U^2$-ONe$ \rm t$. We use well-known Flownet2~\cite{ilg2017flownet} and Mask-RCNN~\cite{he2017mask} respectively to obtain the results of optical flow and instance segmentation as inputs. Before the optical flow is fed through the network, we do the normalization to highlight the moving objects more. The $ \rm U^2$-ONe$ \rm t$ is build with our Octave ReSidual U-block (ORSU) based on Octave Convolution and multiscale attention mechanism based on the convolutional block attention module (CBAM)~\cite{woo2018cbam}. Inspired by Octave~\cite{chen2019drop} and $ \rm U^2$-Ne$ \rm t$~\cite{qin2020u2}, Octave U-block (ORSU) is designed to capture intra-stage multi-scale features while reducing the redundancy issue and computation cost in CNNs. For the motion segmentation map obtained from $ \rm U^2$-ONe$ \rm t$, we use post-processing to combine it with the instance segmentation results to obtain instance-level moving object segmentation results. We extract the contours of the motion segmentation map, and use each closed motion contour to determine whether each semantically instance mask is moving and find new moving instances.

\subsection{ORSU-blocks}
\begin{figure}[t]
	\begin{center}
		\includegraphics[width=1.0\linewidth]{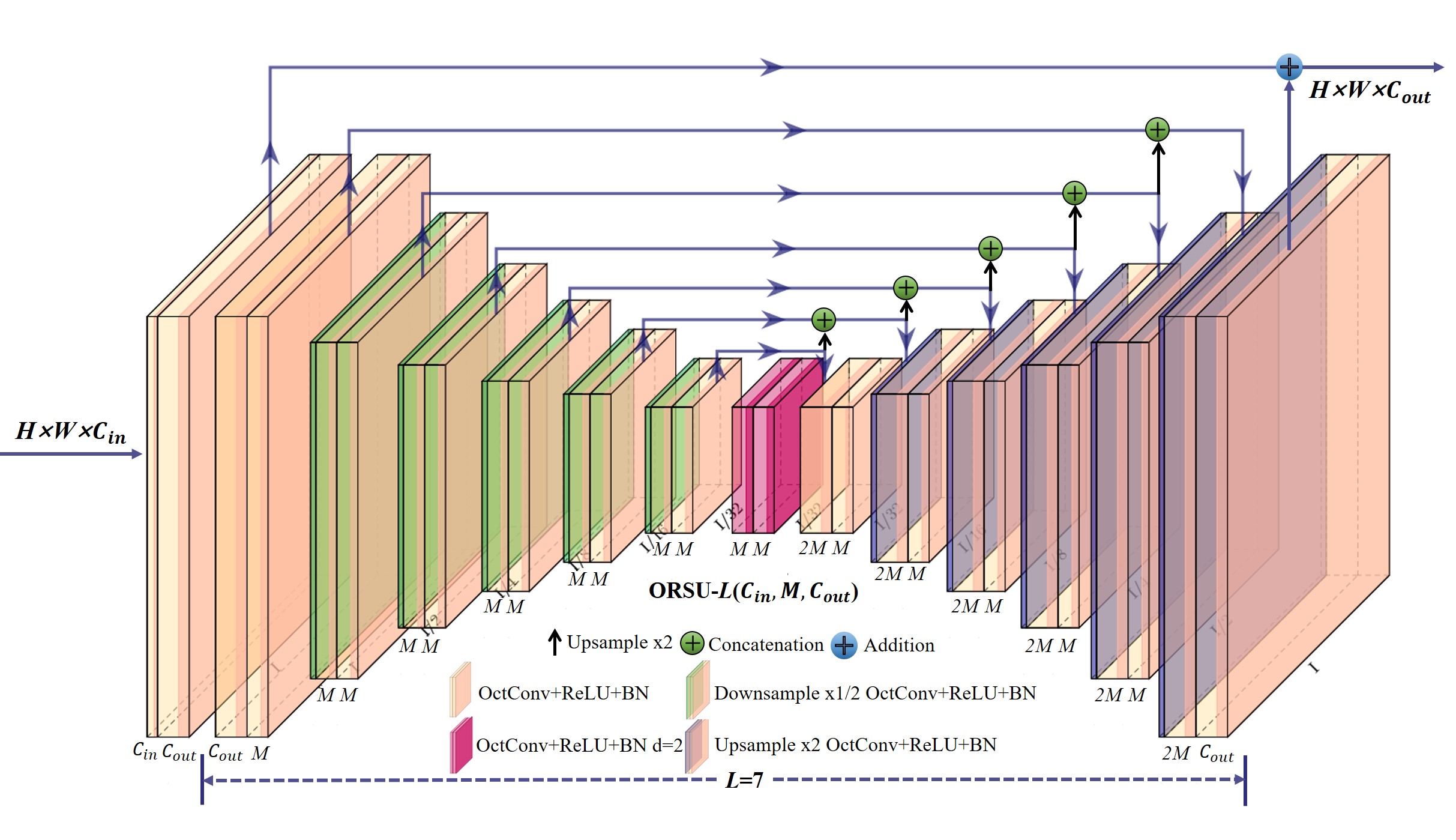}
	\end{center}
	\caption{Our designed Octave ReSidual U-block (ORSU). $L$ is the number of encoder layers, $C_{in}$, $C_{out}$ denote input and output channels, and $M$ is the number of channels in the inner layer of ORSU.}
	\label{fig:fig2}
\end{figure}

Inspired by $ \rm U^2$-Ne$ \rm t$~\cite{qin2020u2}, we propose a novel Octave ReSidual U-block (ORSU) in order to make good use of both local and global contextual information for improving the segmentation effect. ORSU-$\emph L$($C_{in}$,$M$,$C_{out}$) shown in Fig.~\ref{fig:fig2} follows the main structure of RSU in $ \rm U^2$-Ne$ \rm t$. Therefore, our ORSU is also mainly composed of three parts:

(i) an input convolution layer, which uses Octave Convolution (OctConv) for local feature extraction instead of the vanilla convolution. Compared with RSU, ORSU using OctConv further reduces computation and memory consumption while boosting accuracy for segmentation. This layer transforms the input feature map $X(H \times W \times C_{in})$ to an intermediate map $F_1(x)$ with channel of $C_{out}$. 

(ii) a U-Net like symmetric encoder-decoder structure with height of $L$ which is deeper with larger $L$. It takes $F_1(x)$ from the input convolution layer as input and learns to extract and encode the multi-scale contextual information $\mu(F_1(x))$, where $\mu$ denotes the U-Net like structure as shown in Fig.~\ref{fig:fig2}.

(iii) a residual connection of fusing local features and the multi-scale features through the summation of: $F_1(x) + \mu(F_1(x))$.

\begin{figure}[t]
	\begin{center}
		\includegraphics[width=1.0\linewidth]{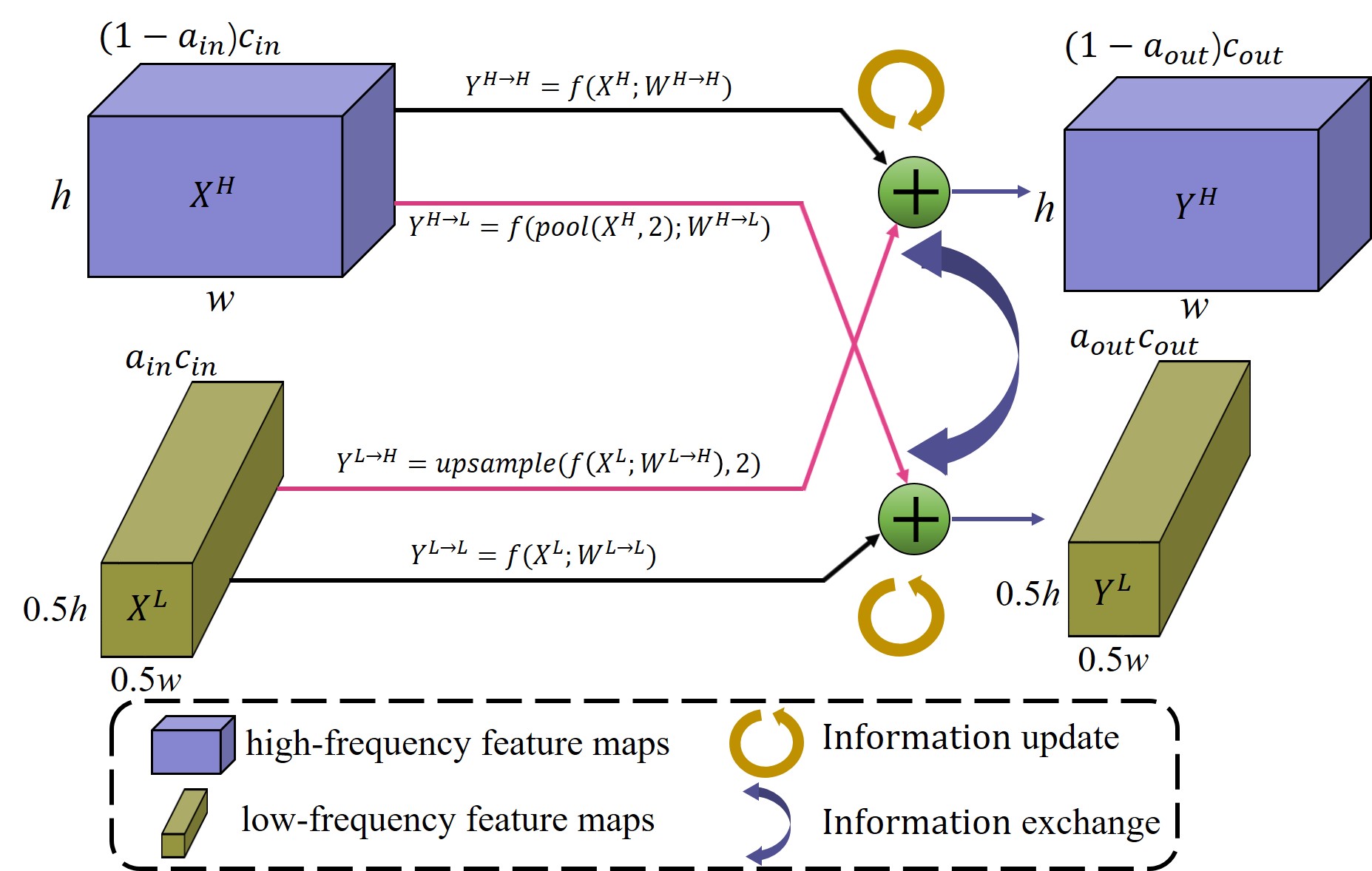}
	\end{center}
	\caption{Illustrating the concept of the Octave Convolution~\cite{chen2019drop}. $\emph f(X; W)$ denotes the convolution function with weight parameters W, $\emph {pool}(X, 2)$ indicates spatial average pooling with kernel size $2 \times 2$ and stride 2, $\emph {upsample}(X)$ indicates an up-sampling operation by a factor of 2.}
	\label{fig:fig3}
\end{figure}

Like RSU, ORSU can capture intra-stage multi-scale features without degradation of high resolution features. The main design difference between our ORSU and RSU is that ORSU replaces vanilla convolutions with Octave Convolution (OctConv). Convolutional neural networks (CNN) have achieved outstanding achievements in many computer vision tasks. However, behind the high accuracy, there is a lot of spatial redundancy that cannot be ignored~\cite{chen2019drop}. Just like the decomposition of spatial frequency components of natural images, OctConv decomposes the output feature maps of a convolution layer into high- and low-frequency feature maps stored in different groups (see Fig.~\ref{fig:fig3}). Therefore, through the information sharing between neighboring locations, the spatial resolution of the low-frequency group can be safely reduced and the spatial redundancy can be reduced. In addition, with the advantage of OctConv performing corresponding (low-frequency) convolution on low-frequency information, we effectively enlarge the receptive field in the pixel space. Therefore, our designed change using OctConv empowers the network to further improve segmentation effect and reduce the computation and memory overhead while retaining the designed advantages of RSU. We show the computation cost comparison between our ORSU and RSU in TABLE.~\ref{Tab01}.

\begin{table}[h]	
	\scriptsize	
	\centering	
	\caption{Comparison of computational cost for blocks and networks}	
	\label{Tab01}
	{\begin{tabular}[l]{@{}lcccccc}	
			\toprule	
			Blocks & FLOPS & Memory & MAdd & \\		
			\midrule	
			RSU-7  & 4.39GFLOPS & 138.67MB & 8.74MAdd  \\	
			ORSU-7 & $\textbf{2.41GFLOPS}$ & $\textbf{123.19MB}$ & $\textbf{4.79MAdd}$  \\
			\midrule	
			Networks & FLOPS & Memory & MAdd & \\
			\midrule
			$ \rm U^2$-Ne$ \rm t$  & 37.67GFLOPS & 444.25MB & 75.20MAdd  \\	
			$ \rm U^2$-ONe$ \rm t$ & $\textbf{21.88GFLOPS}$ & $\textbf{418.56MB}$ & $\textbf{43.64MAdd}$  \\		
			\bottomrule		
	\end{tabular}}
	\label{symbols}
	\vspace{0.2cm}
	\footnotesize{\\
		The results of both blocks and networks are calculated based on an input feature map of dimension $256 \times 256 \times 3$. \\
		The dimension of output feature map of blocks is $256 \times 256 \times 64$.\\
		FLOPS shows theoretical amount of floating point arithmetics, \\
		GFLOPS is Giga Floating Point Operations.\\
		Memory denotes memory usage.\\
		MAdd is theoretical amount of multiply-adds.\\}
\end{table}

\subsection{$ U^2$-ONe$ t$}
Inspired by $ \rm U^2$-Ne$ \rm t$~\cite{qin2020u2}, we propose a noval $ \rm U^2$-ONe$ \rm t$ whose exponential notation 2 is level of nested U-structure. As shown in Fig.~\ref{fig:fig1}, each stage of $ \rm U^2$-ONe$ \rm t$ is filled by a well configured ORSU and there are 11 stages to form a big U-structure. In general, $ \rm U^2$-ONe$ \rm t$ mainly consists of three parts:

(i) A six stages encoder, detailed configurations are presented in TABLE.~\ref{Tab02}. The number "$L$" behind the “ORSU-” denote the height of blocks. $C_{in}$, $M$ and $C_{out}$ represent input channels, middle channels and output channels of each block respectively. The larger $L$ is used to capture more large scale information of the feature map with larger height and width. In both $En\_5$ and $En\_6$ stages, ORSU-4F are used which is dilated version of the ORSU using dilated convolutions (see Fig.~\ref{fig:fig1}) because the resolution of feature maps in these two stages are relatively low. 

(ii) A five stages decoder has similar structures to their symmetrical encoder stages (see Fig.~\ref{fig:fig1} and TABLE.~\ref{Tab02}). The concatenation of the upsampling feature map of the previous stage and the upsampling feature map of the symmetric encoder stage is input for each decoder stage.

\begin{figure}[t]
	\begin{center}
		\includegraphics[width=1.0\linewidth]{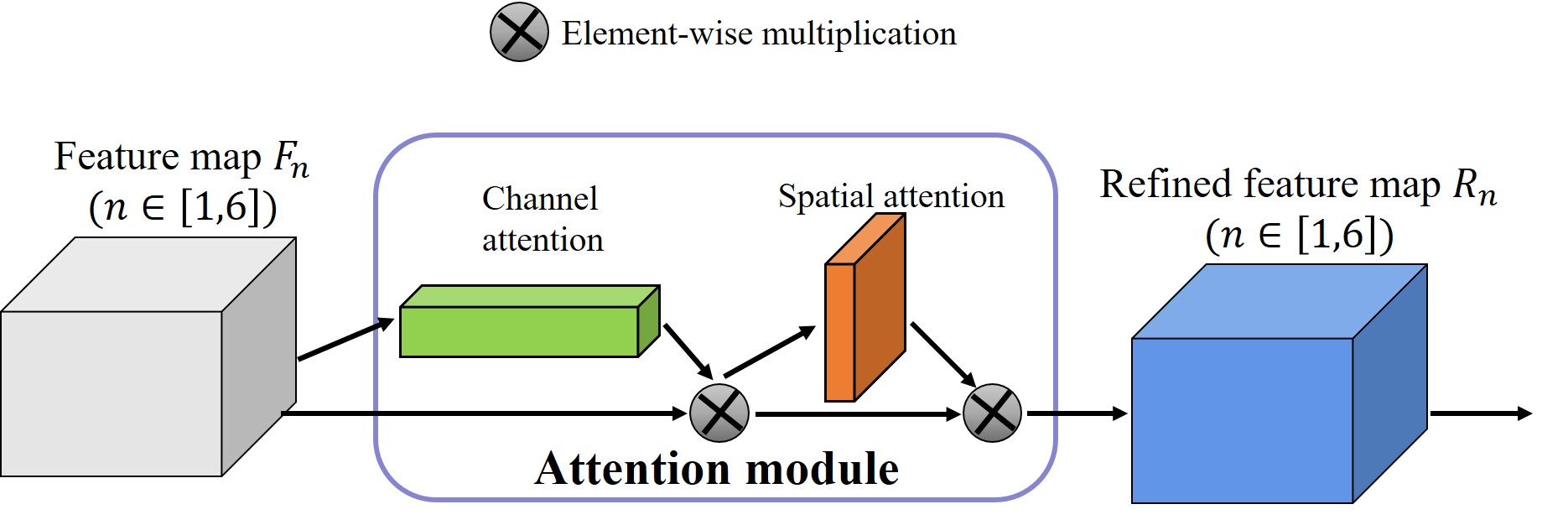}
	\end{center}
	\caption{The overview of the convolutional block attention module (CBAM)~\cite{woo2018cbam}.}
	\label{fig:fig4}
\end{figure}

(iii) The last part is a multiscale attention mechanisms attached with the decoder stages and the last encoder stage. At each level of the network, we add an attention module including channel and spatial attention mechanism inspired by~\cite{woo2018cbam} (see Fig.~\ref{fig:fig4}) to eliminate the aliasing effect that should be eliminated by 3x3 convolution. At the same time, through the channel attention mechanism to assign different significance to the channels of the feature map and the spatial attention mechanism to discover which parts of the feature map are more important, the saliency of the spatial dimension of the moving objects is enhanced. Compared to $ \rm U^2$-Ne$ \rm t$, we maintain a deep architecture with high resolution while further improving segmentation accuracy and reducing the computational and memory cost (see TABLE.~\ref{Tab01} and TABLE.~\ref{Tab03-1}).

\begin{table}[h]	
	\scriptsize	
	\centering	
	\caption{Detailed configuration of the ORSU block at each stage}	
	\hspace{4cm}
	\label{Tab02}
	\setlength{\tabcolsep}{1.6mm}
	{
		\begin{tabular}{ccccccc}
			\toprule
			\multirow{1}{*}{} & \multicolumn{6}{c}{Stages} \\
			\cmidrule(r){2-7} 
			&$\rm En\_1$ &$\rm En\_2$ &$\rm En\_3$ &$\rm En\_4$ &$\rm En\_5$ &$\rm En\_6$ \\
			\midrule
			Block             &ORSU-7   &ORSU-6    &ORSU-5   &ORSU-4 &ORSU-4F &ORSU-4F\\
			$C_{in}$            &3   &64    &128   &256  &512 &512\\
			$M$               &32   &32    &64  &128  &256 &256\\
			$C_{out}$           &64   &128    &256  &512  &512 &512\\
			\midrule
			&$\rm De\_5$ &$\rm De\_4$ &$\rm De\_3$ &$\rm De\_2$ &$\rm De\_1$ \\
			\midrule
			Block             &ORSU-4F   &ORSU-4    &ORSU-5   &ORSU-6 &ORSU-7 \\
			$C_{in}$            &1024   &1024    &512   &256 &128 \\
			$M$               &256   &128    &64  &32 &16\\
			$C_{out}$           &512   &256    &128  &64 &64\\
			\bottomrule
		\end{tabular}
	}
\end{table}

\subsection{Training Supervision Strategy}
Generally speaking, the standard top-most supervised training is not a problem for relatively shallow networks. However, for extremely deep networks, the networks will slow down or not converge or converge to a local optimum due to the gradient disappearance problem during gradient back propagation. Deeply Supervised Networks (DSN)~\cite{lee2015deeply} is proposed to alleviate the optimization difficulties caused by gradient flows through long chains. However, it is still susceptible to problems including the interference of the hierarchical representation generation process and the inconsistency of optimization goals. In the training process, we design a hierarchical training supervision strategy instead of using the standard top-most supervised training and the deep supervision scheme. For each level, we use both the standard binary cross-entropy (BCEloss) and Kullback-Leibler divergence (KLloss) inspired in~\cite{li2020dynamic} to calculate the loss. By adding a pairwise probability prediction matching loss (KLloss) between any two levels, we promote multi-level interaction between different levels. The optimization objectives of the losses in different levels are consistent, thus ensuring the robustness and generalization performance of the model. The ablation study in Section~\ref{sec:ablation on training supervision} proves the effectiveness of our hierarchical training supervision strategy.
The binary cross-entropy loss is:
\begin{equation}\label{eq1}
\resizebox{1.0\hsize}{!}{$l_{BCE}=-\sum\limits_{(i,j)}^{(M,N)}{[G_{(i,j)}\log{S_{(i,j)}}+(1-G_{(i,j)})\log{(1-S_{(i,j)})}]},$}
\end{equation}
and the Kullback-Leibler divergence loss is:
\begin{equation}\label{eq2}
l_{KL}=\sum\limits_{(i,j)}^{(M,N)}{G_{(i,j)}\log{\frac{G_{(i,j)}}{S_{(i,j)}}}}
\end{equation}
where (i, j) is the pixel coordinates and (M, N) is the height and width of the image. $G_{(i,j)}$ and $S_{(i,j)}$ denote the pixel values of the ground truth and the predicted moving object segmentation result respectively.
Our training loss is defined as:
\begin{equation}\label{eq3}
\zeta=\sum\limits_{h=1}^H{w_{BCE}^{(h)}l_{BCE}^{(h)}}+\sum\limits_{h=1}^H{w_{KL}^{(h)}l_{KL}^{(h)}}
\end{equation}
where $l_{BCE}^{(h)}$ and $l_{KL}^{(h)}$ (M = 6, as the Sup1, Sup2, · · · , Sup6 in Fig.~\ref{fig:fig1}) denote the binary cross-entropy loss and the Kullback-Leibler divergence loss of the side output moving object segmentation result $S_{side}^{(h)}$ respectively. $w_{BCE}^{(h)}$ and $w_{KL}^{(h)}$ are the weights of each loss term. We try to minimize the overall loss $\zeta$ of Eq\ref{eq3} in the training process. In the testing process, we choose the $S_{side}^{(h)}$ (M=1) as our final moving object segmentation result.

\subsection{Post-processing}
Through the output of the network, we can obtain the result of moving object segmentation, which is that the foreground moving object region and background are separated, but the instance-level moving object segmentation result is not obtained. In order to get instance-level results, we fuse the semantic instance label mask from Mask-RCNN with the contour extraction results of the motion segmentation map. The geometric contour of the motion segmentation map can improve the quality of the semantic instance mask boundaries, determine whether the instance object is moving, and find new moving objects which it is not associated with a particular semantic class. Meanwhile, the semantic instance mask can provide the category label of some moving objects and boundaries to distinguish overlapping objects for the motion segmentation map. 

Our contour extraction method follows this approach~\cite{suzuki1985topological}, which utilizes topological structural analysis of digitized binary images to obtain multiple closed contours of the motion segmentation map. For each motion contour $C_i$, we calculate the overlap of each semantic instance mask $m_j$ and $C_i$, to associate $m_j$ and $C_i$. Only if this overlap is greater than a threshold – in our experiments $80\%\cdot|m_j|$, where $\cdot|m_j|$ denotes the number of pixels belonging to the mask $m_j$ – $m_j$ is associated with $C_i$. Finally, we obtain the instance-level moving object segmentation result and segment new objects according to the number of semantic instance masks associated with each $C_i$ (see Algorithm.~\ref{alg:Framwork}).

\begin{algorithm}[htb] 
	\caption{ The instance-level moving object segmentation}  
	\label{alg:Framwork}  
	\begin{algorithmic}[1]  
		\Require  
		Each motion contour $C_i$;
		Each instance semantic mask $m_j$;
		Results of semantic instance masks associated with $C_i$; 
		Judgment threshold $t$ ($t$ usually take 200 in our experiments);
		\Ensure  
		Each moving object instance and its mask;  
		\For{each motion contour $C_i$ $\in$ current motion segmentation map}  
		\If {the number of semantic instance mask associated with the motion contour $C_i$ $>$ 1}  
		\For{each semantic instance mask $m_j$ associated with the motion contour $C_i$}  
		\State $m_j$ is output as the mask for a moving object instance
		\State the number of moving object instance $\gets$ the number of moving object instance + 1.
		\EndFor 
		\EndIf
		\If {the number of semantic instance mask associated with the motion contour $C_i$ $==$ 1}
		\State The area contained in motion contour $C_i$ is assigned as the mask for associated semantic instance $j$.
		\State $m_j$ is output as the mask for a moving object instance.
		\State the number of moving object instance $\gets$ the number of moving object instance + 1.
		\EndIf 
		\If {the number of semantic instance mask associated with the motion contour $C_i$ $<$ 1}
		\If {the length of the motion contour $C_i$ $>$ $t$}
		\State The area contained in motion contour $C_i$ is output as the mask for a new moving object instance.
		\State the number of moving object instance $\gets$ the number of moving object instance + 1.
		\EndIf  
		\EndIf  
		\EndFor  
	\end{algorithmic}  
\end{algorithm}  

\section{Experiments}
\begin{figure*}
	\begin{center}
		\includegraphics[width=1.0\linewidth]{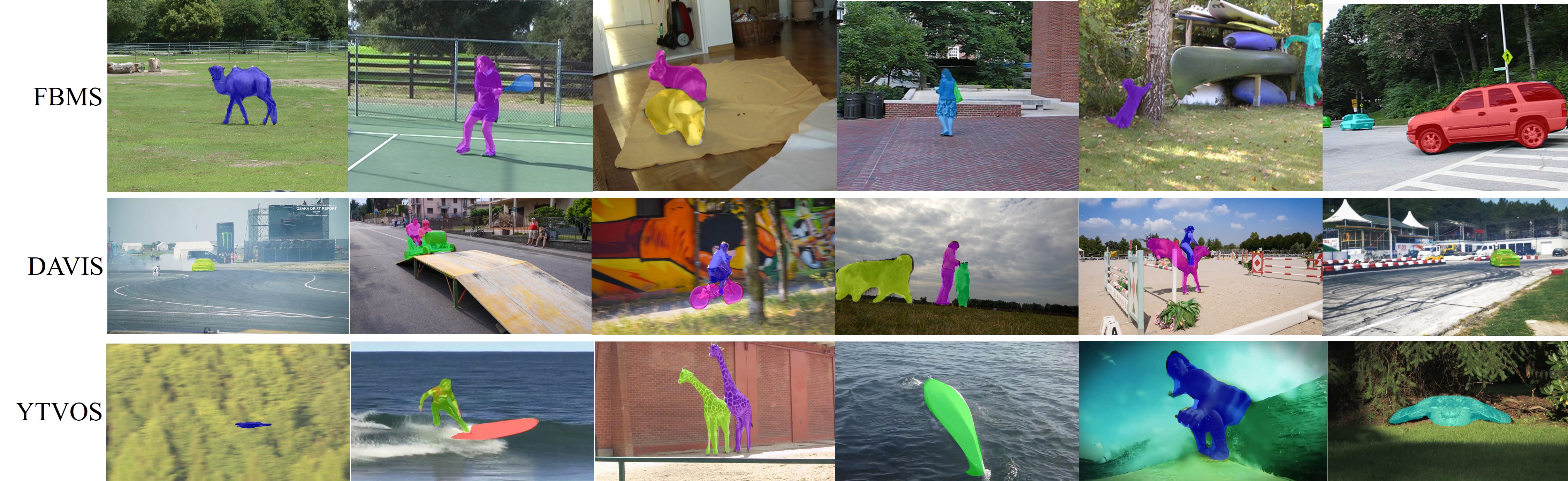}
	\end{center}
	\caption{Qualitative results for three datasets.}
	\label{fig:fig8}
\end{figure*}

\textbf{Datasets.} We evaluate our method on several frequently used benchmark datasets: FBMS~\cite{ochs2013segmentation}, DAVIS~\cite{perazzi2016benchmark,pont20172017}, YouTube Video Object Segmentation (YTVOS)~\cite{xu2018youtube}. For FBMS, we evaluate on the Testset using the model trained from Trainingset. FBMS shows a large number of annotation errors. We use a corrected version of the dataset linked to the original data set's website~\cite{bideau2016detailed}. For DAVIS, DAVIS16~\cite{perazzi2016benchmark} is comprised of 50 sequences containing instance segmentation masks for only the moving objects. Unlike DAVIS 2016, the DAVIS2017~\cite{pont20172017} contains sequences providing instance-level masks for both moving and static objects, not all of its sequences are suitable for our model. Therefore, we train our model on DAVIS16 and use a subset of the DAVIS 17 called DAVIS-Moving defined in~\cite{dave2019towards} for evaluation. For YTVOS containing both labeled static or moving objects, we also use the YTVOS-Moving dataset introduced by~\cite{dave2019towards} which selects sequences where all moving objects are labeled. Qualitative results for our method are shown in Fig.~\ref{fig:fig8}.

\textbf{Implementation Details.} We train our network from scratch and all of convolutional layers are initialized by~\cite{he2015delving}. The stochastic gradient descent (SGD) with a initial learning rate of 4e-2 is used for optimization and its hyper parameters settings are as follows: momentum=0.9, weight\_decay=0.0001. We trained 20 epochs using the batch size of 4. Both training and testing are conducted on a single NVIDIA Tesla V100 GPU with 16 GB memory, along with the PyTorch 1.1.0 and Python 3.7 deep learning frameworks. We use precision(P), recall(R), F-measure(F) as defined in~\cite{ochs2013segmentation} and mean intersection over union (IoU) for the evaluation metrics. Source code and the models will be made public available upon the publish of the paper.

\subsection{Ablation Studies}

\begin{table*}[]	
	\centering	
	\caption{Results of ablation studies on blocks and attention mechanism}	
	\label{Tab03-1}
	\begin{tabular}{ccccccccc}
		\toprule
		\multirow{2}{*}{} & \multicolumn{4}{c}{Video Foreground Segmentation} & \multicolumn{4}{c}{Multi-object Motion Segmentation} \\
		\cmidrule(r){2-5} \cmidrule(r){6-9}
		&  P (Precision)      &  R (Recall)   &   F (F-measure) & IoU
		&  P (Precision)      &  R (Recall)   &   F (F-measure) & IoU  \\
		\midrule
		$ \rm U^2$-Ne$ \rm t_{6bk}$(-a)             &84.6748   &78.7881    &80.2232   &70.9541   &79.8170  &74.4494  &75.7438   &70.3260 \\
		\midrule
		$ \rm U^2$-ONe$ \rm t_{6bk}$(-a)             &87.0821   &79.7173    &81.6738   &72.4611   &82.3621  &75.5477  &77.3492   &72.1066 \\
		\midrule
		$ \rm U^2$-ONe$ \rm t_{6bk}$           &88.3126   &82.5591    &83.7723   &74.1656   &83.1169  &77.9250  &78.9910  &72.9785 \\
		\bottomrule
	\end{tabular}
	\vspace{0.2cm}
	\footnotesize{
	\\ Video Foreground Segmentation denotes segmenting the scene into static background or foreground moving objects without distinguishing object instances.
    \\ Multi-object Motion Segmentation is segmenting each moving object instance in scene.}
\end{table*}

\begin{table*}[]	
	\centering	
	\caption{Results of ablation study on training supervision}	
	\label{Tab03-2}
	\begin{tabular}{ccccccccc}
		\toprule
		\multirow{2}{*}{} & \multicolumn{4}{c}{Video Foreground Segmentation} & \multicolumn{4}{c}{Multi-object Motion Segmentation} \\
		\cmidrule(r){2-5} \cmidrule(r){6-9}
		&  P (Precision)      &  R (Recall)   &   F (F-measure) & IoU
		&  P (Precision)      &  R (Recall)   &   F (F-measure) & IoU  \\
		\midrule
		$ \rm U^2$-ONe$ \rm t_{1bk}$            &87.1959   &82.9170    &83.3599   &73.7483   &82.0288  &78.2415  &78.5725   &72.7251 \\
		$ \rm U^2$-ONe$ \rm t_{1b}$            &87.6808   &82.4263    &83.2078   &73.4435   &82.4819  &77.7830  &78.4259   &72.9288 \\
		\midrule
		$ \rm U^2$-ONe$ \rm t_{2bk}$             &86.5249   &$\textcolor[rgb]{1,0,0}{83.6013}$   &83.2855   &73.5262   &81.3052  &$\textcolor[rgb]{1,0,0}{78.8389}$  &78.4271   &73.2597 \\
		$ \rm U^2$-ONe$ \rm t_{2b}$            &87.5558   &$\textcolor[rgb]{0,0,1}{83.3884}$    &83.7176   &74.0066   &82.2579  &$\textcolor[rgb]{0,0,1}{78.6416}$  &78.8341   &73.2570 \\
		\midrule
		$ \rm U^2$-ONe$ \rm t_{3bk}$            &88.0516   &82.7291    &83.6525   &74.0434   &82.9008  &78.0600  &78.9054   &73.2325 \\
		$ \rm U^2$-ONe$ \rm t_{3b}$            &88.2510   &82.3962    &83.5610   &73.9233   &83.0433  &77.7766  &78.7818   &73.2416 \\
		\midrule
		$ \rm U^2$-ONe$ \rm t_{4bk}$            &88.1412   &83.1088    &$\textcolor[rgb]{1,0,0}{83.9566}$   &$\textcolor[rgb]{1,0,0}{74.4435}$  &82.9590  &78.4306 &$\textcolor[rgb]{1,0,0}{79.1666}$  &$\textcolor[rgb]{0,0,1}{73.2620}$ \\
		$ \rm U^2$-ONe$ \rm t_{4b}$            &88.2951   &82.4503    &83.7305   &74.1254   &$\textcolor[rgb]{0,0,1}{83.1775}$  &77.8654  &79.0048   &73.1279 \\
		\midrule
		$ \rm U^2$-ONe$ \rm t_{5bk}$           &88.2397   &82.2916    &83.4473   &73.7114   &83.0693  &77.6426  &78.6916   &72.8023 \\
		$ \rm U^2$-ONe$ \rm t_{5b}$            &$\textcolor[rgb]{1,0,0}{88.8535}$   &82.2117    &83.7729   &$\textcolor[rgb]{0,0,1}{74.2275}$   &$\textcolor[rgb]{1,0,0}{83.6419}$  &77.6087  &79.0023  &72.7234 \\
		\midrule
		$ \rm U^2$-ONe$ \rm t_{6bk}$           &$\textcolor[rgb]{0,0,1}{88.3126}$   &82.5591    &83.7723   &74.1656  &83.1169  &77.9250  &78.9910  &72.9785 \\
		$ \rm U^2$-ONe$ \rm t_{6b}$            &88.1056   &82.9051    &$\textcolor[rgb]{0,0,1}{83.8449}$   &74.1089   &82.9020  &78.2278  &$\textcolor[rgb]{0,0,1}{79.0357}$   &$\textcolor[rgb]{1,0,0}{73.3622}$ \\
		\bottomrule
	\end{tabular}
	\vspace{0.2cm}
	\footnotesize{
		\\ Best results are highlighted in $\rm \textcolor[rgb]{1,0,0}{red}$ with second best in $\rm \textcolor[rgb]{0,0,1}{blue}$.}
\end{table*}

\subsubsection{Ablation on Blocks}
The blocks ablation is to verify that the ORSU structure of our design improves the accuracy while reducing the computational and memory cost, compared with RSU. We removed the attention mechanism in our network and then only replace ORSU block with RSU block designed by $ \rm U^2$-Net to get the network called $ \rm U^2$-Ne$ \rm t_{6bk}$(-a). $ \rm U^2$-Ne$ \rm t_{6bk}$(-a) is a network for motion segmentation that uses the backbone of $ \rm U^2$-Net and calculates the BCEloss and KLloss of six level as we designed. The results are shown in TABLE.~\ref{Tab01} and TABLE.~\ref{Tab03-1}, after just replacing RSU with our ORSU, memory usage drop by 25.69MB and computational cost fall by nearly 40\%. For both video foreground segmentation and multi-object motion segmentation, the network with ORSU improves the precision by over 2.4\%, the recall by about 1.0\%, F-measure by over 1.55\% and IoU by over 1.51\%. At the same time, we notice that the increasing value of the evaluation metrics in multi-object motion segmentation is higher than that in video foreground segmentation. It is worth noting that the improvement of precision is the most obvious, indicating that our ORSU helps the network to better learn motion information and more accurately segment moving objects. Therefore, we believe that our designed ORSU is better than RSU in motion segmentation tasks. 

\subsubsection{Ablation on Attention Mechanism}
As we mentioned above, the addition of the attention mechanism introduces spatial and channel attention, making the moving object more salient in the segmentation result. This ablation is to validate the effectiveness of adding attention mechanism. We compare $ \rm U^2$-ONe$ \rm t_{6bk}$(-a) without the attention mechanism and our complete network called $ \rm U^2$-ONe$ \rm t_{6bk}$. TABLE.~\ref{Tab03-1} shows that adding attention mechanisms improves the precision by about 1\%, the recall by over 2.37\%, F-measure by over 1.64\% and IoU by over 0.87\%. Different from the blocks ablation, the increasing value of the evaluation metrics after adding attention mechanism in video foreground segmentation is higher than that in multi-object motion segmentation. Meanwhile we notice that the improvement of recall is the most obvious, indicating that designed multiscale attention mechanism promotes the network to better discover more moving objects. This is done by introducing global context information and capturing the spatial details around moving object to enhance the saliency of the spatial dimension of the moving objects. 

\subsubsection{Ablation on Training Supervision}
\label{sec:ablation on training supervision}
In order to prove the effectiveness of our multi-level loss calculation strategy, we evaluate the network $ \rm U^2$-ONe$ \rm t_{bk}$ when calculating the BCEloss and KLloss of from 1 to 6 levels separately. TABLE.~\ref{Tab03-2} shows that the overall effect of the network calculating the loss of from 2 to 4 levels separately is improving as the number of levels increases. The overall effect using from 4 to 6 levels appears to decline first and then increase, as the number of levels increases. In summary, we believe that our multi-level loss calculation training strategy further improves the effect of our deep network, but it is not that the more levels, the better the effect. For our network, it probably works best when using six-levels or four-levels.

To further demonstrate the superiority of the calculation of both BCEloss and KLloss, we conduct multiple ablation experiments. From 1 to 6 levels separately, we compared the network only calculating BCEloss called $ \rm U^2$-ONe$ \rm t_{b}$ and the network calculating both BCEloss and KLloss called $ \rm U^2$-ONe$ \rm t_{bk}$. The result is shown in TABLE.~\ref{Tab03-2}. From the results of calculating the loss of 3 and 4 levels, the addition of KLloss improves the effect of the network in general although improvement of each metric is not obvious unlike the previous ablation studies. When calculating the loss of greater than 4 levels, the addition of KLloss seems to slightly reduce the performance of the network. We think this is mainly because the negative impact of the loss calculation level above 4 levels causes the decrease in the effect of the network. And the interaction between the various levels brought by KLloss further increases this negative impact. Overall, the addition of KLloss improves the performance of our network, so we obtained the best network we think – $ \rm U^2$-ONe$ \rm t_{4bk}$, when calculating both the BCEloss and KLloss  of 4 levels. At the same time, we find that the addition of KLloss improves the precision and makes the network more accurately segment moving objects, which is what we need for some practical applications. 

\subsection{Comparison to Prior Work}
\textbf{Official FBMS.} We evaluate our method against prior work on the standard FBMS Testset using the model trained from FBMS Trainingset. The input image size is (512, 640). Since some of the methods we compare only provide metrics of multi-object motion segmentation task, we only compare the metrics of this task. In order to compare with other methods to indicate the accuracy of the segmented object count, we add $\Delta$Obj metric as defined in~\cite{xie2019object}. As shown in TABLE.~\ref{Tab04}, our model performs the best in recall. In terms of recall and F-measure, we outperform greatly CCG~\cite{bideau2018best}, OBV~\cite{xie2019object} and STB~\cite{shen2018submodular} respectively by over 16.5\% and over 6.4\%. Qualitative results are shown in Fig.~\ref{fig:fig8}.

\begin{table}[h]	
	\centering	
	\caption{FBMS results using the official metric}	
	\label{Tab04}
	\begin{tabular}{cccccc}
		\toprule
		\multirow{1}{*}{} & \multicolumn{4}{c}{Multi-object Motion Segmentation} \\
		\cmidrule(r){2-6} 
		&  P      &  R   &   F & IoU & $\Delta$Obj \\
		\midrule
		CCG~\cite{bideau2018best}             &74.23   &63.07   &64.97   &--   &4 \\
		OBV~\cite{xie2019object}             &75.90   &66.60    &67.30   &-- &4.9 \\
		STB~\cite{shen2018submodular}        &$\textcolor[rgb]{0,0,1}{87.11}$   &66.53    &75.44   &-- &-- \\
		TSA~\cite{dave2019towards}            &$\textcolor[rgb]{1,0,0}{88.60}$   &$\textcolor[rgb]{0,0,1}{80.40}$    &$\textcolor[rgb]{1,0,0}{84.30}$  &--  &-- \\
		Ours            &84.80   &$\textcolor[rgb]{1,0,0}{83.10}$    &$\textcolor[rgb]{0,0,1}{81.84}$  &79.70 &4.9  \\
		\bottomrule
	\end{tabular}
	\hspace{4cm}
	\footnotesize{
		\\ Best results are highlighted in $\rm \textcolor[rgb]{1,0,0}{red}$ with second best in $\rm \textcolor[rgb]{0,0,1}{blue}$.}
\end{table}

\textbf{DAVIS and YTVOS.} We further evaluate our work on DAVIS-Moving and evaluation set of YTVOS-Moving defined in~\cite{dave2019towards}. The input image size of both datasets is (512, 896). The results is shown in TABLE.~\ref{Tab05}. In DAVIS-Moving, our approach outperforms TSA~\cite{dave2019towards} by about 4.8\% in precision and about 0.8\% in F-measure. Unlike FBMS and DAVIS, YTVOS-Moving contains many moving objects that are difficult to segment, such as snakes, octopuses, camouflaged objects. Therefore, the metrics in YTVOS-Moving are strongly lower than the metrics of the previous datasets. However, in YTVOS-Moving, our method still outperforms TSA by about 4.2\% in recall and about 1.6\% in F-measure. Qualitative results are shown in Fig.~\ref{fig:fig8}.

\begin{table}[h]
	\scriptsize		
	\centering	
	\caption{Results for DAVIS-Moving and YTVOS-Moving defined in~\cite{dave2019towards}}	
	\hspace{4cm}
	\label{Tab05}
	\begin{tabular}{ccccccc}	
		\hline	
		\multicolumn{7}{c}{Multi-object Motion Segmentation} \\
		\cmidrule(r){2-7}
		Dataset & & P & R & F & IoU \\	
		\hline	
		\multirow{2}{*}{DAVIS-Moving}  &TSA~\cite{dave2019towards} & 78.30		
		&$\textbf{78.80}$	&78.10 &-- \\		
		\cline{2-6}	
		& Ours & $\textbf{83.12}$	
		& 77.93 & $\textbf{78.99}$ & 72.98	
		&   \\		
		\cline{1-6}		
		\multirow{2}{*}{YTVOS-Moving} &TSA~\cite{dave2019towards} & 74.50	
		& 66.40	&68.30 &--  \\		
		\cline{2-6}		
		& Ours & $\textbf{74.64}$		
		& $\textbf{70.56}$ & $\textbf{69.93}$ & 65.67	
		&  \\		
		\cline{1-6}		
		\hline		
	\end{tabular}		
\end{table}

\begin{figure}[t]
	\centering
	\subfigure[] {
		\label{fig:fig6-a}     
		\includegraphics[width=0.313\linewidth]{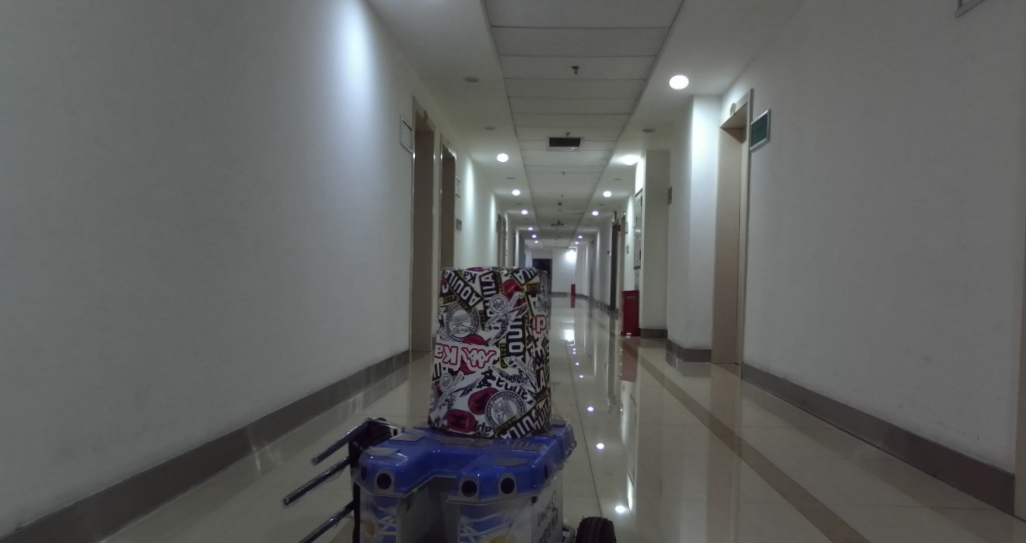}  
	} 
	\hspace{-0.4cm}
	\subfigure[] {
		\label{fig:fig6-b}     
		\includegraphics[width=0.313\linewidth]{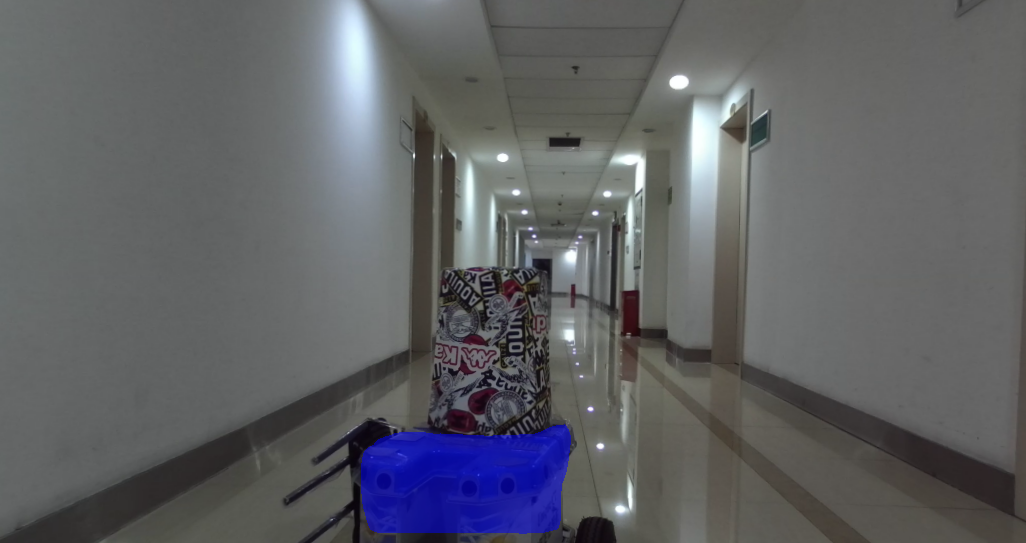}  
	} 
	\hspace{-0.4cm}
	\subfigure[] {
		\label{fig:fig6-c}     
		\includegraphics[width=0.313\linewidth]{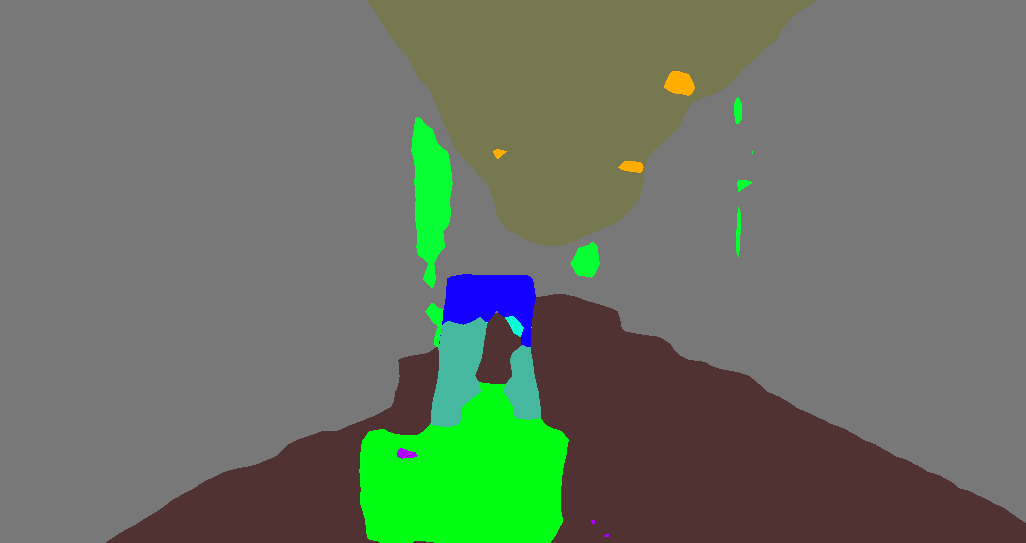}  
	} 
	\caption{(a) The mobile robot without semantic labels in our dataset. (b) Instance segmentation result from Mask-RCNN. (c) Semantic segmentation result from PSPnet~\cite{zhao2017pyramid}. Neither method can segment this object well.}
	\label{fig:fig6}
\end{figure}

\begin{figure}[t]
	\centering
	\subfigure[] {
		\label{fig:fig7-a}     
		\includegraphics[width=0.31\linewidth]{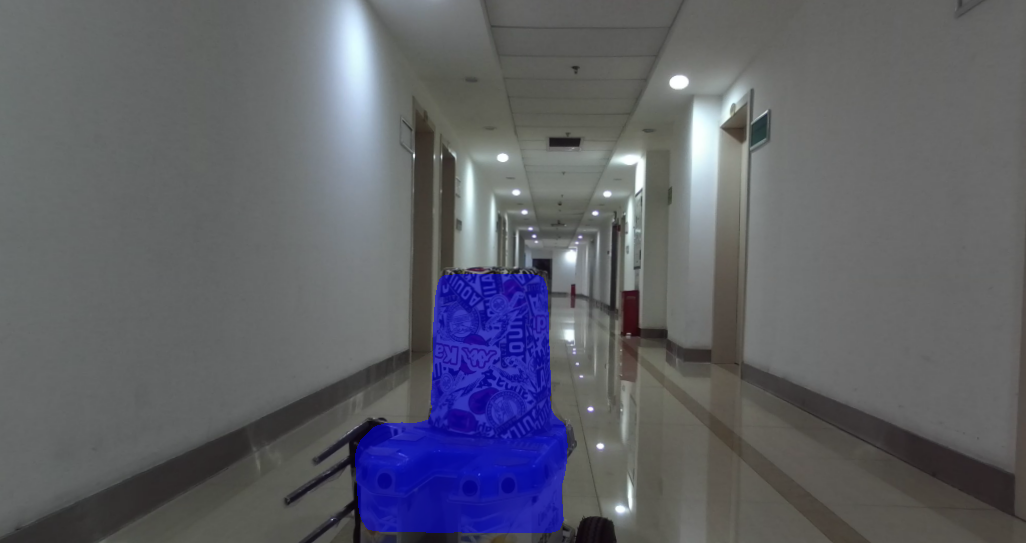}  
	} 
	\hspace{-0.4cm}
	\subfigure[] {
		\label{fig:fig7-b}     
		\includegraphics[width=0.31\linewidth]{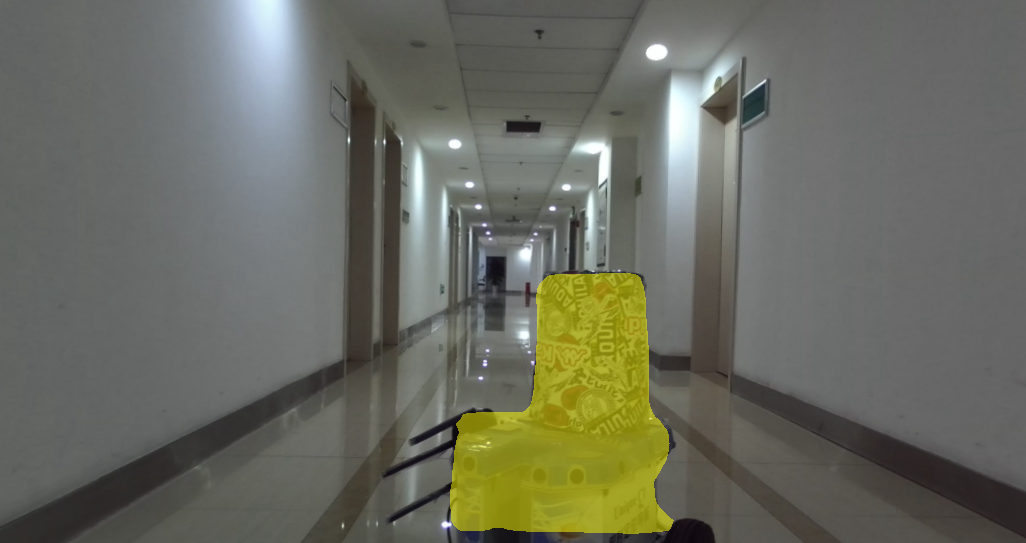}  
	} 
	\hspace{-0.4cm}
	\subfigure[] {
		\label{fig:fig7-c}     
		\includegraphics[width=0.31\linewidth]{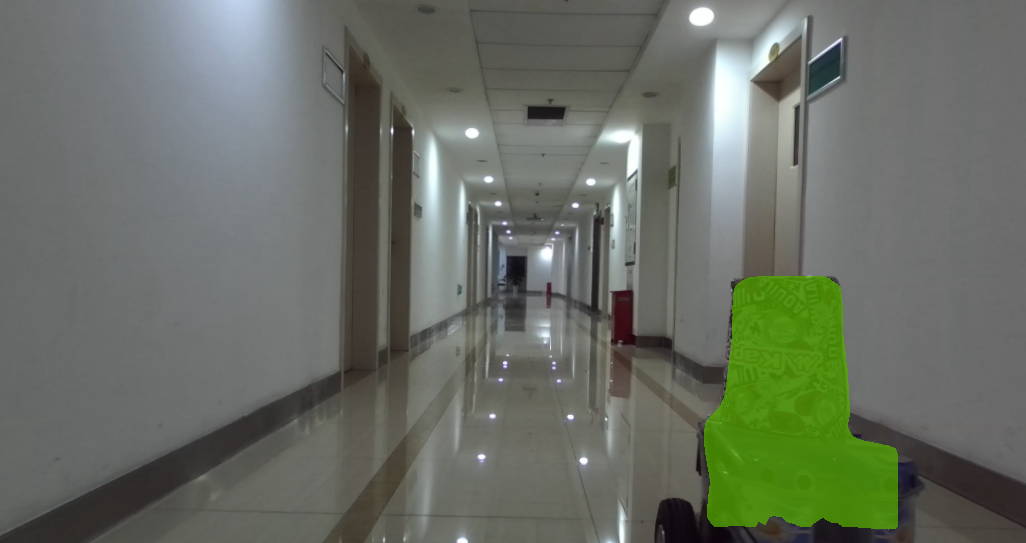}  
	} 
	\caption{Qualitative results for our method in this dataset.}
	\label{fig:fig7}
\end{figure}

\textbf{Our dataset.} In order to prove that our method has the ability to discover new objects without predefined semantic labels, we conduct qualitative experiments in our own datasets. There is always a mobile robot without semantic labels in our dataset (see Fig~\ref{fig:fig6}). We use the model trained from DAVIS dataset for tesing, without any training in our dataset. Qualitative results are shown in Fig.~\ref{fig:fig7}. Our method can segment this moving new object, proving the efficiency of our method.

\section{Conclusion}
In this paper, we propose a two-level nested U-structure with multiscale attention mechanism called $ \rm U^2$-ONe$ \rm t$ for moving instances segmentation. Each stage of $ \rm U^2$-ONe$ \rm t$ is filled with our new designed Octave ReSidual U-block (ORSU) based on Octave Convolution, which enables $ \rm U^2$-ONe$ \rm t$ to capture rich both local and global information with high resolution while reducing the spatial redundancy and computation resources in CNNs. And we design a hierarchical training supervision strategy which adds KLloss to improve the consistency of the whole deep network optimization and calculates both BCEloss and KLloss at all six levels for improving the effectiveness of our deep network. Experimental results on several general moving objects segmentation datasets show that our work is the state-of-the-art method. Especially in some challenging datasets, such as YTVOS-Moving including camouflaged objects and tiny objects, our method still achieves good performance. Experiments in our dataset also prove that our work has the ability to segment new objects. 

In the near future, we will further leverage the anti-noise ability of Octave Convolution to combine with the ability to extract multi-scale features of ORSU. And we will introduce the network for image dehazing and deraining to enhance the images before being fed through the network. Finally, we enable $ \rm U^2$-ONe$ \rm t$ to still achieve good performance under extremely complex conditions including rain, fog, snow, motion blur, and containing a lot of noise, etc.


%

\ifCLASSOPTIONcaptionsoff
  \newpage
\fi



\bibliographystyle{IEEEtran}
\bibliography{IEEEabrv,egbib}

\begin{thebibliography}{10}
\providecommand{\url}[1]{#1}
\csname url@samestyle\endcsname
\providecommand{\newblock}{\relax}
\providecommand{\bibinfo}[2]{#2}
\providecommand{\BIBentrySTDinterwordspacing}{\spaceskip=0pt\relax}
\providecommand{\BIBentryALTinterwordstretchfactor}{4}
\providecommand{\BIBentryALTinterwordspacing}{\spaceskip=\fontdimen2\font plus
\BIBentryALTinterwordstretchfactor\fontdimen3\font minus
  \fontdimen4\font\relax}
\providecommand{\BIBforeignlanguage}[2]{{%
\expandafter\ifx\csname l@#1\endcsname\relax
\typeout{** WARNING: IEEEtran.bst: No hyphenation pattern has been}%
\typeout{** loaded for the language `#1'. Using the pattern for}%
\typeout{** the default language instead.}%
\else
\language=\csname l@#1\endcsname
\fi
#2}}
\providecommand{\BIBdecl}{\relax}
\BIBdecl

\bibitem{saputra2018visual}
M.~R.~U. Saputra, A.~Markham, and N.~Trigoni, ``Visual slam and structure from
  motion in dynamic environments: A survey,'' \emph{ACM Computing Surveys
  (CSUR)}, vol.~51, no.~2, p.~37, 2018.

\bibitem{zhao2019motion}
X.~Zhao, Q.~Qin, and B.~Luo, ``Motion segmentation based on model selection in
  permutation space for rgb sensors,'' \emph{Sensors}, vol.~19, no.~13, p.
  2936, 2019.

\bibitem{zhang2017permutation}
Y.~Zhang, B.~Luo, and L.~Zhang, ``Permutation preference based alternate
  sampling and clustering for motion segmentation,'' \emph{IEEE Signal
  Processing Letters}, vol.~25, no.~3, pp. 432--436, 2017.

\bibitem{he2017mask}
K.~He, G.~Gkioxari, P.~Doll{\'a}r, and R.~Girshick, ``Mask r-cnn,'' in
  \emph{Proceedings of the IEEE international conference on computer vision},
  2017, pp. 2961--2969.

\bibitem{badrinarayanan2017segnet}
V.~Badrinarayanan, A.~Kendall, and R.~Cipolla, ``Segnet: A deep convolutional
  encoder-decoder architecture for image segmentation,'' \emph{IEEE
  transactions on pattern analysis and machine intelligence}, vol.~39, no.~12,
  pp. 2481--2495, 2017.

\bibitem{ronneberger2015u}
O.~Ronneberger, P.~Fischer, and T.~Brox, ``U-net: Convolutional networks for
  biomedical image segmentation,'' in \emph{International Conference on Medical
  image computing and computer-assisted intervention}.\hskip 1em plus 0.5em
  minus 0.4em\relax Springer, 2015, pp. 234--241.

\bibitem{redmon2018yolov3}
J.~Redmon and A.~Farhadi, ``Yolov3: An incremental improvement,'' \emph{arXiv
  preprint arXiv:1804.02767}, 2018.

\bibitem{bideau2018best}
P.~Bideau, A.~RoyChowdhury, R.~R. Menon, and E.~Learned-Miller, ``The best of
  both worlds: Combining cnns and geometric constraints for hierarchical motion
  segmentation,'' in \emph{Proceedings of the IEEE Conference on Computer
  Vision and Pattern Recognition}, 2018, pp. 508--517.

\bibitem{xie2019object}
C.~Xie, Y.~Xiang, Z.~Harchaoui, and D.~Fox, ``Object discovery in videos as
  foreground motion clustering,'' in \emph{Proceedings of the IEEE Conference
  on Computer Vision and Pattern Recognition}, 2019, pp. 9994--10\,003.

\bibitem{dave2019towards}
A.~Dave, P.~Tokmakov, and D.~Ramanan, ``Towards segmenting anything that
  moves,'' in \emph{Proceedings of the IEEE International Conference on
  Computer Vision Workshops}, 2019, pp. 0--0.

\bibitem{muthu2020motion}
S.~Muthu, R.~Tennakoon, T.~Rathnayake, R.~Hoseinnezhad, D.~Suter, and
  A.~Bab-Hadiashar, ``Motion segmentation of rgb-d sequences: Combining
  semantic and motion information using statistical inference,'' \emph{IEEE
  Transactions on Image Processing}, vol.~29, pp. 5557--5570, 2020.

\bibitem{qin2020u2}
X.~Qin, Z.~Zhang, C.~Huang, M.~Dehghan, O.~R. Zaiane, and M.~Jagersand,
  ``U2-net: Going deeper with nested u-structure for salient object
  detection,'' \emph{Pattern Recognition}, vol. 106, p. 107404, 2020.

\bibitem{chen2019drop}
Y.~Chen, H.~Fan, B.~Xu, Z.~Yan, Y.~Kalantidis, M.~Rohrbach, S.~Yan, and
  J.~Feng, ``Drop an octave: Reducing spatial redundancy in convolutional
  neural networks with octave convolution,'' in \emph{Proceedings of the IEEE
  International Conference on Computer Vision}, 2019, pp. 3435--3444.

\bibitem{faktor2014video}
A.~Faktor and M.~Irani, ``Video segmentation by non-local consensus voting.''
  in \emph{BMVC}, vol.~2, no.~7, 2014, p.~8.

\bibitem{wang2015saliency}
W.~Wang, J.~Shen, and F.~Porikli, ``Saliency-aware geodesic video object
  segmentation,'' in \emph{Proceedings of the IEEE conference on computer
  vision and pattern recognition}, 2015, pp. 3395--3402.

\bibitem{perazzi2016benchmark}
F.~Perazzi, J.~Pont-Tuset, B.~McWilliams, L.~Van~Gool, M.~Gross, and
  A.~Sorkine-Hornung, ``A benchmark dataset and evaluation methodology for
  video object segmentation,'' in \emph{Proceedings of the IEEE Conference on
  Computer Vision and Pattern Recognition}, 2016, pp. 724--732.

\bibitem{wang2019learning}
W.~Wang, H.~Song, S.~Zhao, J.~Shen, S.~Zhao, S.~C. Hoi, and H.~Ling, ``Learning
  unsupervised video object segmentation through visual attention,'' in
  \emph{Proceedings of the IEEE conference on computer vision and pattern
  recognition}, 2019, pp. 3064--3074.

\bibitem{wang2019zero}
W.~Wang, X.~Lu, J.~Shen, D.~J. Crandall, and L.~Shao, ``Zero-shot video object
  segmentation via attentive graph neural networks,'' in \emph{Proceedings of
  the IEEE international conference on computer vision}, 2019, pp. 9236--9245.

\bibitem{lu2019see}
X.~Lu, W.~Wang, C.~Ma, J.~Shen, L.~Shao, and F.~Porikli, ``See more, know more:
  Unsupervised video object segmentation with co-attention siamese networks,''
  in \emph{Proceedings of the IEEE conference on computer vision and pattern
  recognition}, 2019, pp. 3623--3632.

\bibitem{yang2019anchor}
Z.~Yang, Q.~Wang, L.~Bertinetto, W.~Hu, S.~Bai, and P.~H. Torr, ``Anchor
  diffusion for unsupervised video object segmentation,'' in \emph{Proceedings
  of the IEEE international conference on computer vision}, 2019, pp. 931--940.

\bibitem{zhuo2019unsupervised}
T.~Zhuo, Z.~Cheng, P.~Zhang, Y.~Wong, and M.~Kankanhalli, ``Unsupervised online
  video object segmentation with motion property understanding,'' \emph{IEEE
  Transactions on Image Processing}, vol.~29, pp. 237--249, 2019.

\bibitem{yang2020collaborative}
Z.~Yang, Y.~Wei, and Y.~Yang, ``Collaborative video object segmentation by
  foreground-background integration,'' \emph{arXiv preprint arXiv:2003.08333},
  2020.

\bibitem{girshick2014rich}
R.~Girshick, J.~Donahue, T.~Darrell, and J.~Malik, ``Rich feature hierarchies
  for accurate object detection and semantic segmentation,'' in
  \emph{Proceedings of the IEEE conference on computer vision and pattern
  recognition}, 2014, pp. 580--587.

\bibitem{ren2015faster}
S.~Ren, K.~He, R.~Girshick, and J.~Sun, ``Faster r-cnn: Towards real-time
  object detection with region proposal networks,'' in \emph{Advances in neural
  information processing systems}, 2015, pp. 91--99.

\bibitem{liu2018path}
S.~Liu, L.~Qi, H.~Qin, J.~Shi, and J.~Jia, ``Path aggregation network for
  instance segmentation,'' in \emph{Proceedings of the IEEE conference on
  computer vision and pattern recognition}, 2018, pp. 8759--8768.

\bibitem{kang2020bshapenet}
B.~R. Kang, H.~Lee, K.~Park, H.~Ryu, and H.~Y. Kim, ``Bshapenet: Object
  detection and instance segmentation with bounding shape masks,''
  \emph{Pattern Recognition Letters}, vol. 131, pp. 449--455, 2020.

\bibitem{peng2020deep}
S.~Peng, W.~Jiang, H.~Pi, X.~Li, H.~Bao, and X.~Zhou, ``Deep snake for
  real-time instance segmentation,'' in \emph{Proceedings of the IEEE/CVF
  Conference on Computer Vision and Pattern Recognition}, 2020, pp. 8533--8542.

\bibitem{hurtik2020poly}
P.~Hurtik, V.~Molek, J.~Hula, M.~Vajgl, P.~Vlasanek, and T.~Nejezchleba,
  ``Poly-yolo: higher speed, more precise detection and instance segmentation
  for yolov3,'' \emph{arXiv preprint arXiv:2005.13243}, 2020.

\bibitem{xu20193d}
X.~Xu, L.~F. Cheong, and Z.~Li, ``3d rigid motion segmentation with mixed and
  unknown number of models,'' \emph{IEEE Transactions on Pattern Analysis and
  Machine Intelligence}, 2019.

\bibitem{thakoor2010multibody}
N.~Thakoor, J.~Gao, and V.~Devarajan, ``Multibody structure-and-motion
  segmentation by branch-and-bound model selection,'' \emph{IEEE Transactions
  on Image Processing}, vol.~19, no.~6, pp. 1393--1402, 2010.

\bibitem{ranjan2019competitive}
A.~Ranjan, V.~Jampani, L.~Balles, K.~Kim, D.~Sun, J.~Wulff, and M.~J. Black,
  ``Competitive collaboration: Joint unsupervised learning of depth, camera
  motion, optical flow and motion segmentation,'' in \emph{Proceedings of the
  IEEE conference on computer vision and pattern recognition}, 2019, pp.
  12\,240--12\,249.

\bibitem{pan2019joint}
L.~Pan, Y.~Dai, M.~Liu, F.~Porikli, and Q.~Pan, ``Joint stereo video
  deblurring, scene flow estimation and moving object segmentation,''
  \emph{IEEE Transactions on Image Processing}, vol.~29, pp. 1748--1761, 2019.

\bibitem{shen2018submodular}
J.~Shen, J.~Peng, and L.~Shao, ``Submodular trajectories for better motion
  segmentation in videos,'' \emph{IEEE Transactions on Image Processing},
  vol.~27, no.~6, pp. 2688--2700, 2018.

\bibitem{ilg2017flownet}
E.~Ilg, N.~Mayer, T.~Saikia, M.~Keuper, A.~Dosovitskiy, and T.~Brox, ``Flownet
  2.0: Evolution of optical flow estimation with deep networks,'' in
  \emph{Proceedings of the IEEE conference on computer vision and pattern
  recognition}, 2017, pp. 2462--2470.

\bibitem{woo2018cbam}
S.~Woo, J.~Park, J.-Y. Lee, and I.~So~Kweon, ``Cbam: Convolutional block
  attention module,'' in \emph{Proceedings of the European conference on
  computer vision (ECCV)}, 2018, pp. 3--19.

\bibitem{lee2015deeply}
C.-Y. Lee, S.~Xie, P.~Gallagher, Z.~Zhang, and Z.~Tu, ``Deeply-supervised
  nets,'' in \emph{Artificial intelligence and statistics}, 2015, pp. 562--570.

\bibitem{li2020dynamic}
D.~Li and Q.~Chen, ``Dynamic hierarchical mimicking towards consistent
  optimization objectives,'' in \emph{Proceedings of the IEEE/CVF Conference on
  Computer Vision and Pattern Recognition}, 2020, pp. 7642--7651.

\bibitem{suzuki1985topological}
S.~Suzuki \emph{et~al.}, ``Topological structural analysis of digitized binary
  images by border following,'' \emph{Computer vision, graphics, and image
  processing}, vol.~30, no.~1, pp. 32--46, 1985.

\bibitem{ochs2013segmentation}
P.~Ochs, J.~Malik, and T.~Brox, ``Segmentation of moving objects by long term
  video analysis,'' \emph{IEEE transactions on pattern analysis and machine
  intelligence}, vol.~36, no.~6, pp. 1187--1200, 2013.

\bibitem{pont20172017}
J.~Pont-Tuset, F.~Perazzi, S.~Caelles, P.~Arbel{\'a}ez, A.~Sorkine-Hornung, and
  L.~Van~Gool, ``The 2017 davis challenge on video object segmentation,''
  \emph{arXiv preprint arXiv:1704.00675}, 2017.

\bibitem{xu2018youtube}
N.~Xu, L.~Yang, Y.~Fan, D.~Yue, Y.~Liang, J.~Yang, and T.~Huang, ``Youtube-vos:
  A large-scale video object segmentation benchmark,'' \emph{arXiv preprint
  arXiv:1809.03327}, 2018.

\bibitem{bideau2016detailed}
P.~Bideau and E.~Learned-Miller, ``A detailed rubric for motion segmentation,''
  \emph{arXiv preprint arXiv:1610.10033}, 2016.

\bibitem{he2015delving}
K.~He, X.~Zhang, S.~Ren, and J.~Sun, ``Delving deep into rectifiers: Surpassing
  human-level performance on imagenet classification,'' in \emph{Proceedings of
  the IEEE international conference on computer vision}, 2015, pp. 1026--1034.

\bibitem{zhao2017pyramid}
H.~Zhao, J.~Shi, X.~Qi, X.~Wang, and J.~Jia, ``Pyramid scene parsing network,''
  in \emph{Proceedings of the IEEE conference on computer vision and pattern
  recognition}, 2017, pp. 2881--2890.

\end{thebibliography}
%

%






\end{document}